\documentclass[10pt,twocolumn,letterpaper]{article}
\usepackage[pagenumbers]{cvpr}
\usepackage{algorithm}
\usepackage{multirow}
\usepackage[table]{xcolor}
\usepackage{amsmath,amssymb,amsfonts}%
\usepackage[normalem]{ulem}
\useunder{\uline}{\ul}{}
\usepackage{makecell}
\usepackage{comment} 
\usepackage{stfloats}

\newcommand{\wangxi}[1]{{\color{black}#1}}  

\newcommand{\frameworkName}{\textcolor{black}{LineArt }}
\newcommand{\datasetName}{\textcolor{black}{ProLines }}

\definecolor{cvprblue}{rgb}{0.21,0.49,0.74}
\usepackage[pagebackref,breaklinks,colorlinks,allcolors=cvprblue]{hyperref}

\title{LineArt: A Knowledge-guided Training-free High-quality Appearance Transfer for Design Drawing with Diffusion Model}

\author{Xi Wang$^\ddagger$ 
\hspace{.9cm} Hongzhen Li$^\ddagger$ \hspace{.9cm} Heng Fang$^\uplus$ 
\hspace{.9cm} Yichen Peng$^\amalg$
\\ Haoran Xie$^\diamond$ 
\hspace{.7cm} Xi Yang$^\ddagger$\thanks{Corresponding author}
\hspace{0.7cm} Chuntao Li$^\ddagger$ 
\\
$^\ddagger$ Jilin University.\\
$^\uplus$ KTH Royal Institute of Technology.\\
$^\amalg$ Tokyo Institute of Technology.\\
$^\diamond$ Japan Advanced Institute of Science and Technology (JAIST). 
}

\begin{document}

\twocolumn[{
\renewcommand\twocolumn[1][]{#1}
\maketitle
\begin{center}
    \captionsetup{type=figure}
    \vspace{-6mm}
    \includegraphics[width=\textwidth]{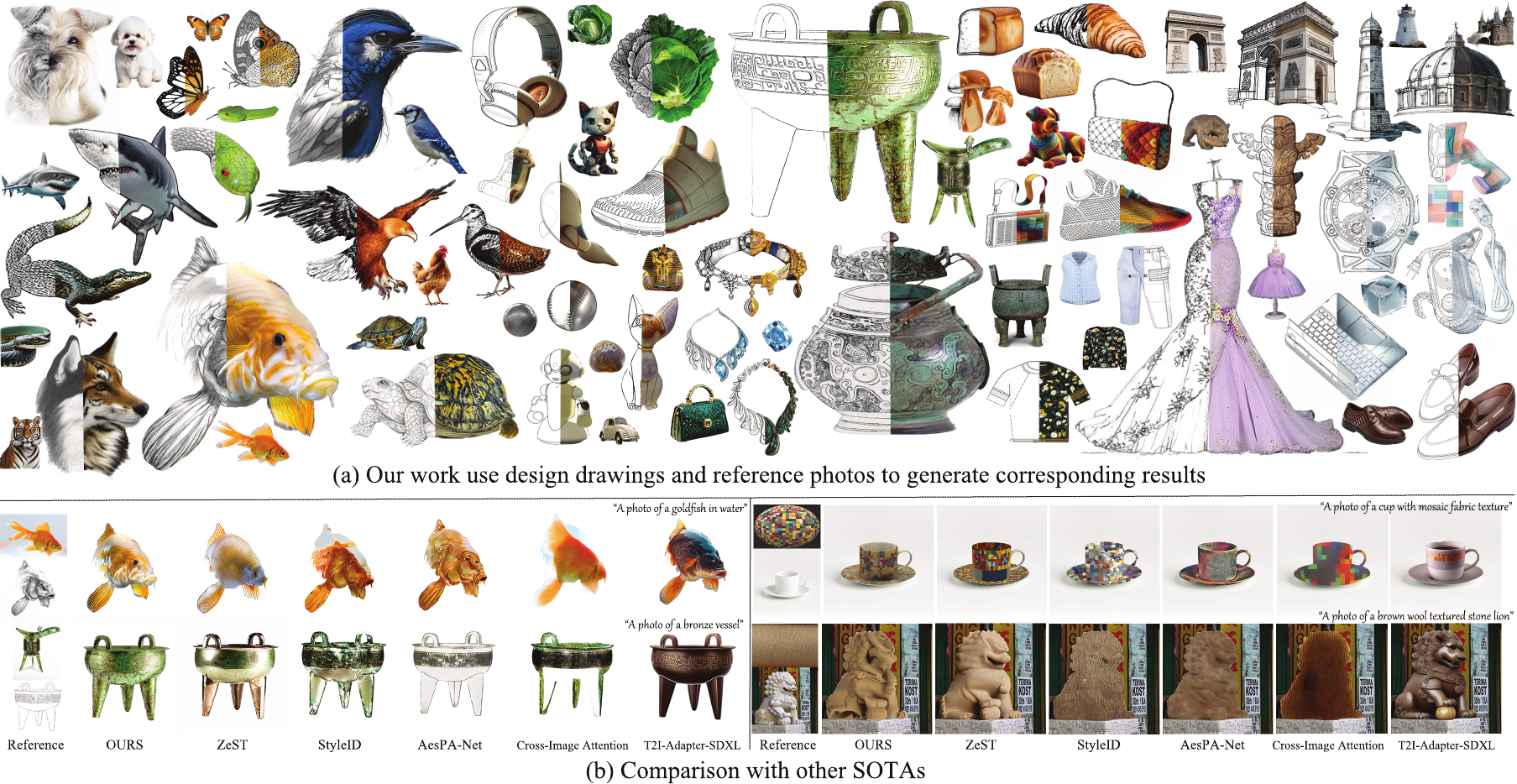}
    \vspace{-7mm}
    \captionof{figure}{\textbf{Generated results by our method and Comparison with SOTAs.}
    (a): Given an object-centered \wangxi{design drawing} and a photo as the reference, our method transfers the complex appearance features in the photo to the fine structure of the design drawing with high fidelity.
    (b): Compared with ZeST~\cite{ZeST}, StyleID~\cite{StyleID}, AesPA-Net~\cite{AesPA-Net}, Cross-Image Attention~\cite{crossimageattention}, and T2I-Adapter-SDXL~\cite{T2I-Adapter}. Our method better ensures the accurate transfer of complex appearance features and maintains the fine structure of the original design drawing.}
    \label{fig:teaser}
\end{center}
}]

\begin{abstract}
\wangxi{Image rendering from line drawings is vital in design and image generation technologies reduce costs, yet professional line drawings demand preserving complex details. Text prompts struggle with accuracy, and image translation struggles with consistency and fine-grained control.} We present \frameworkName, a framework that transfers complex appearance onto detailed design drawings, facilitating design and artistic creation. 
It generates high-fidelity appearance while preserving structural accuracy by simulating hierarchical visual cognition and integrating human artistic experience to guide the diffusion process.
\frameworkName overcomes the limitations of current methods in terms of difficulty in fine-grained control and style degradation in design drawings. It requires no precise 3D modeling, physical property specs, or network training, making it more convenient for design tasks. \frameworkName consists of two stages: a multi-frequency lines fusion module to supplement the input design drawing with detailed structural information and a two-part painting process for Base Layer Shaping and Surface Layer Coloring. 
We also present a new design drawing dataset \datasetName for evaluation. The experiments show that \frameworkName performs better in accuracy, realism, and material precision compared to SOTAs.
\end{abstract}
\section{Introduction}
\label{sec:intro}
Image rendering from line drawing holds significant applications in the fields of design and artistic creation~\cite{yeh2024texturedreamer}. Traditionally, this process requires extensive manual modeling and rendering attempts~\cite{sketchrenderer_2020, Sketch2Mesh_ICCV2021, GET3D_NEURIPS2022, text-based3D_ICCV2023, Sketchdream_TOG2024}. With the rise of image generation technology, it has become possible to control generated attributes through multi-modal inputs such as text and reference images~\cite{text-basedt2P_WACV2022, carrillo_2023CVPR, Masksketch_2023CVPR, yeh2024texturedreamer, sharma2024alchemist, li2023gligen}, or by collecting paired training datasets to achieve domain transfer from line drawings to photographs~\cite{sketchGAN_ICCV2021, sketch2Photo_ECCV2020, COGS_ECCV2022, wang2022unsupervised, chen2018sketchygan}. This advancement has greatly reduced the cost of image rendering from line drawings and significantly accelerated the design iteration process.

However, there remain many challenges in transforming line drawing into photorealistic images through domain transfer. Existing research has largely focused on simple, abstract, user-level hand-drawn sketches~\cite{Sc-fegan_2019CVPR, face_2021, koley_CVPR2023, ItYourSketch_CVPR2024, chen2020deepfacedrawing, gao2020sketchycoco, voynov_2023SigG}, while there is a noticeable gap in the study of \wangxi{professional design drawing} which contains more detailed levels and more complex structural information. Achieving a generation process that preserves these essential details while producing realistic material places higher demands on the performance of image generation models, particularly when working with fine-grained drawing types~\cite{bhunia2020sketch, parmar2023zero, zhang2023prospect}.
Furthermore, high-quality domain data is difficult to collect, making it challenging to rely on large-scale paired training datasets. As shown in the Figure~\ref{fig:teaser}, the text conditioning approach struggles to accurately capture complex appearance and texture requirements~\cite{Dreambooth}. Meanwhile, image translation methods rely on effectively decoupling appearance and structure within image prompts~\cite{huang2017arbitrary, wu2021styleformer, StyleDiffusion_2023ICCV, deng2022stytr2}; otherwise, the generated results often struggle to maintain identity consistency~\cite{faceIdentity_2023ICCV} and lack fine-grained control and editing over multiple attributes~\cite{wei2021fine}. Additionally, conflicts arising from inherent distribution differences between prompts can lead to unavoidable attribute degradation as shown in Figure~\ref{fig:teaser}.

Inspired by human artistic creation, we tackle the above challenges with a knowledge-guided, non-training solution. We develop \frameworkName that leverages widely used ControlNet~\cite{ControlNet} and IP-Adaptor~\cite{IP-Adapter} to transfer complex appearance properties onto \wangxi{design drawings} with fine structures, generating high-fidelity results that align well with appearance in structural details. Our framework has two key stages: the first stage involves decomposing and augmenting the line drawing to better control the structural details of the object; the second extracts brightness and texture features from the appearance prompt to guide the generation of illumination and texture quality in the image.

In the first stage, \frameworkName is inspired by vision representation~\cite{marr1976early, marr1980visual} and incorporates a line-based hierarchical fusion module (multi-frequency lines fusion) that supplements the original input design drawing into an ideal sketch covering three levels of information: continuous single lines for region division~\cite{hertzmann2020line}, double lines to emphasize local details, and a discrete point set of soft edges to represent implicit spatial gradients and texture information.
In the second stage, the input reference image is decomposed to obtain both the brightness control factor and a texture map. Drawing from the coloring process in oil painting, we divide this stage into two parts: Base Layer Shaping and Surface Layer Coloring. In the first part, the brightness control factor simulates the underpainting in oil painting, providing initial light and dark contrasts and shaping structures as a foundation for subsequent lighting effects. In the second part, the texture map undergoes global encoding and is then selectively injected into specific attention layers of the U-net architecture to simulate the translucent layering of colors in oil painting. This division between two parts reduces the complexity of depicting light and shadow directly through color while preserving the original drawing’s structural details and clarity.

In this work we made the following main contributions:
\begin{itemize}
\setlength{\itemsep}{0pt}
\setlength{\parsep}{0pt}
\setlength{\parskip}{0pt}
    \item{we propose a novel method for generating high-quality images from \wangxi{design drawings}. By integrating human painting knowledge and simulating the human process of hierarchical visual information processing, we have developed an innovative, non-training-based method for this task. The relevant design of the method is intuitive and has good understandability and interpretability.}

    \item{We propose a multi-frequency lines fusion module to refine visual cues for better presentation. By decomposing the edge features of the input drawing into three levels of abstraction, \frameworkName achieves detailed structure control. We innovatively split the diffusion denoising into two stages \textit{Base Layer Shaping} and \textit{Surface Layer Coloring} to enhance control over appearance generation.}
    \item{We propose a subdivided generation task for handling drawings, as each line in a sketch serves a different purpose. To support this, we collected a professional design drawing dataset, \datasetName, focused on objects, and established filtering rules and workflows based on a complexity analysis of the drawings. Through extensive comparative experiments, we demonstrate that \frameworkName significantly outperforms various state-of-the-art works, excelling in accuracy, realism, and material precision.} 
\end{itemize}
\section{Related Work}
\noindent \textbf{Sketch-to-Image (S2I) Generation:} S2I generation is a rapidly growing field dedicated to converting sketches into photorealistic images. Early work often views S2I as a domain transfer task, focusing on sparse, abstract user-drawn sketches~\cite{chen2020deepfacedrawing, gao2020sketchycoco}. Traditional GAN-based models optimize generation through techniques such as contextual loss~\cite{lu2018image}, multistage generation~\cite{ghosh2019interactive}, or improve image fidelity by mapping sketches to the latent space of pre-trained GANs~\cite{koley_CVPR2023, richardson2021encoding, Sc-fegan_2019CVPR, sketchGAN_ICCV2021}. Diffusion-based methods have gained popularity, with models like~\cite{wang2022pretraining} using dedicated encoders to map sketches into latent + spaces of pre-trained diffusion models, and SDEdit~\cite{meng2021sdedit} adding noise to sketches, which are then iteratively denoised based on textual prompts. SGDM~\cite{voynov_2023SigG} ensures that noisy images align with sketches, while multi-conditional frameworks like ControlNet~\cite{ControlNet} and T2I-Adapter~\cite{T2I-Adapter} use additional inputs, such as depth maps and color palettes, to improve control over the generated images. Recent models like CoGS~\cite{COGS_ECCV2022} use supervised learning with image-sketch pairs to generate high-quality images even from rough sketches. In contrast, unsupervised approaches like~\cite{Masksketch_2023CVPR} do not require paired training data, allowing for greater flexibility in handling sketches of different abstraction levels and real photos. In our work, we focus specifically on translation tasks that require close consistency with the source image in terms of structural details, ensuring that after image translation, key elements related to the structure remain unchanged, while other aspects are selectively modified. 

\noindent \textbf{Control and Guidance in Diffusion Models:} Diffusion models have gained significant attention as generative models. Researchers in the visual domain have been exploring image-guided stylization based on diffusion models to achieve more personalized generation results~\cite{StyleDiffusion_2023ICCV, StyleID, AesPA-Net}. The study of image-guided stylization can be traced back to the mid-1990s, where early methods used techniques like brush strokes to create stylistic effects. Then, pioneering methods by Neural Style Transfer~\cite{gatys2015neural} and~\cite{ulyanov2016texture} laid the foundation for modern stylization techniques, although these methods were initially limited to a single style. To overcome this limitation,~\cite{ghiasi2017exploring} introduced a style prediction network, enabling a single model to support multiple styles. In 2022,~\cite{deng2022stytr2} demonstrated the superior performance of transformers in stylization tasks, marking a significant advancement in the field.
Since 2023, stylization methods based on diffusion models have garnered increasing attention~\cite{hamazaspyan2023diffusion, qi2024deadiff}. Notable approaches include~\cite{ControlNet}, which uses adapters;~\cite{zhang2023inversion} adopt inversion and shared attention mechanisms; and~\cite{Dreambooth}, a test-time fine-tuning approach. These methods leverage the extensive prior knowledge of diffusion models to interpret and manipulate both structural and artistic elements in images.
\section{Preliminaries}
\begin{figure}[t]
    \centering
    \includegraphics[width=\linewidth]{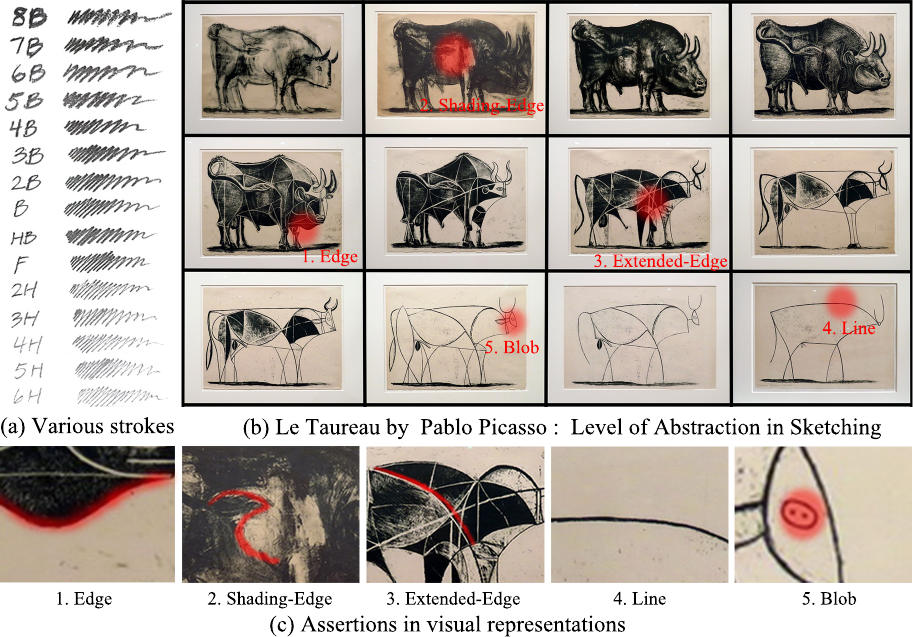}
    \vspace{-6mm}
    \captionof{figure}{
    (a) Various pencils in hardness and color. 
    (b) \textit{Le Taureau} by Picasso depicts a shift from realism to abstraction. ©``Pasadena, Norton Simon Museum, Picasso P. The Bull, 1946'' photo by Vahe Martirosyan, [CC BY-SA 2.0] via \href{https://bit.ly/3MFB3pm}{(https://bit.ly/3MFB3pm)}. 
    (c) Visual representations of \textit{Le Taureau}.}
    \vspace{-6mm}
    \label{fig:3_2_levelOfAbstract}
\end{figure}
\begin{figure*}[t]
    \centering
    \vspace{-6mm}
    \includegraphics[width=\textwidth]{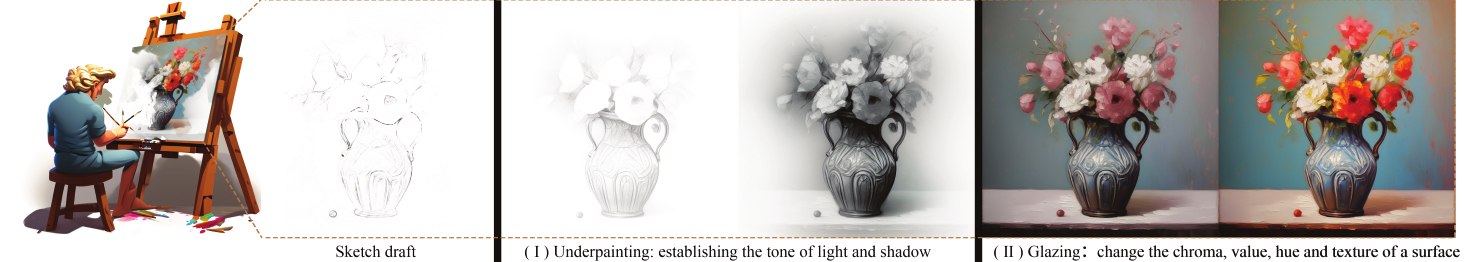}
    \caption{\textbf{A classical painting technique: Imprimatura.} We designate the Underpainting process as Base Layer Shaping and Glazing as Surface Layer Coloring. In the Base Layer Shaping stage, we handle implicit information from the reference image, such as lighting effects, illumination, and shading-based reflectance. Features related to texel and color are addressed in the Surface Layer Coloring stage.}
    \vspace{-6mm}
    \label{fig:3_2_Imprimatura}
\end{figure*}

\begin{figure*}[ht]
    \vspace{-6mm}
    \captionsetup{type=figure}
    \includegraphics[width=\textwidth]{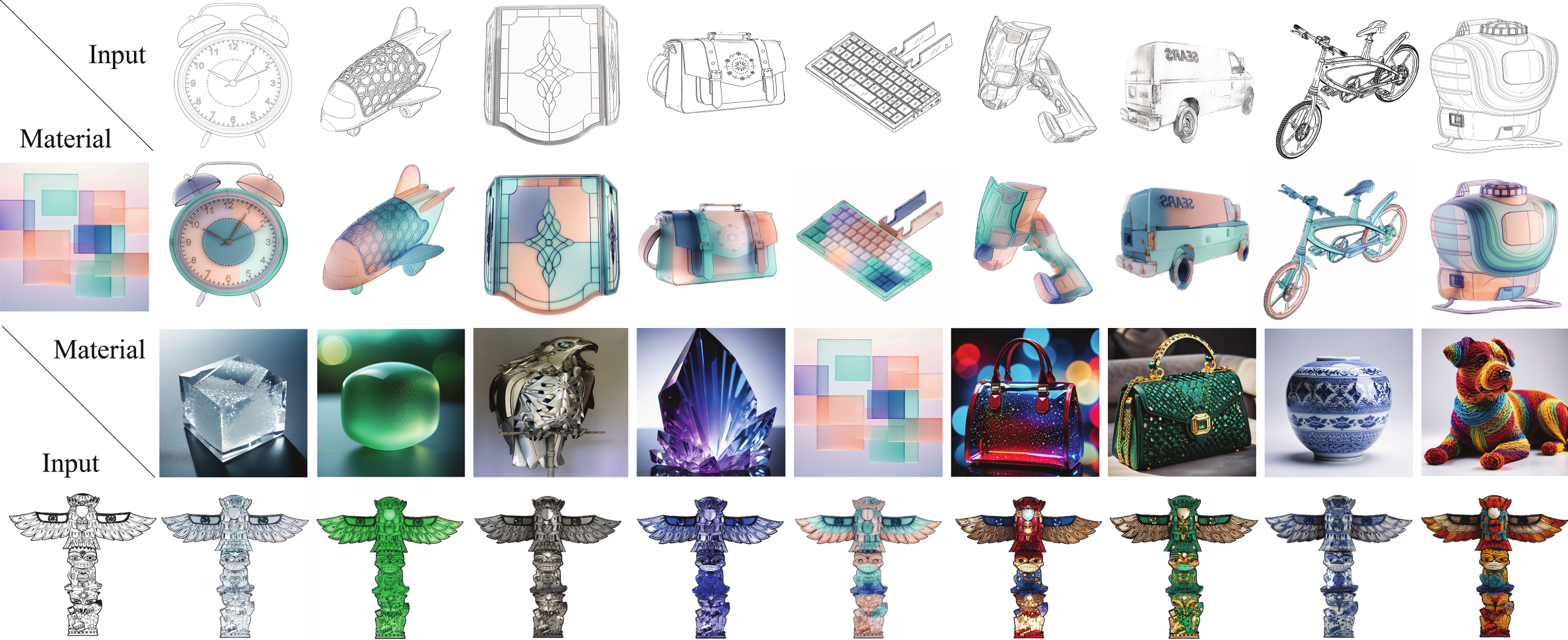}
    \captionof{figure}{\textbf{Examples of our method in assisting design.} The upper part shows that for line draft designers, our method can quickly render the technical drawings with the effect of a specified material, helping designers to intuitively evaluate whether the design of the line draft under the limited material is reasonable. The lower part shows that for fixed product drafts, our method can integrate various material textures with the technical drawings to generate rich material effect previews, assisting designers in choosing the most suitable material combination for product line draft design.}
    \vspace{-3mm}
    \label{fig:designTasks}
\end{figure*}

\noindent \textbf{Problem:} Given a drawing with contours and fine structural details, along with a reference image for appearance texture, the challenge is how to apply complex materials to this object-centered design drawing while preserving structural details and avoiding conflicts with texture features (texels). This issue is common in design tasks (Figure~\ref{fig:designTasks}) and presents three main challenges: (1) maintaining control over structural details while blending diverse textures, (2) appearance features encompass attributes like lighting, texture, and color; degradation in any of these can reduce the design's impact, and (3) it is difficult to collect paired training datasets for design drawings, making it important to explore the model's zero-shot learning capabilities.

\subsection{Assertions: Level of Edge Representations}
\label{section_3.2}
In the field of design drawing, artists need to selectively emphasize key visual features of the subject while downplaying other features to convey the essence of a scene or object~\cite{2023clipascene}. For example in Figure~\ref{fig:3_2_levelOfAbstract}, Picasso's \textit{Le Taureau} series of prints demonstrates a gradual evolution from realism to abstraction~\cite{2022clipasso}. Artists express their visual understanding by manipulating the quantity and complexity of strokes~\cite{bhunia2020pixelor, qu2023sketchxai, cheng2023adaptively}, making it crucial to capture the level of abstraction in the sketch for subsequent generative processes~\cite{bandyopadhyay2024sketchinr, hertzmann2020line}. Algorithms must interpret these information level~\cite{2024sketchmean, tripathi2020sketch} to align with design expectations~\cite{berger2013style, liu2021deflocnet}.

In the early field of visual information processing~\cite{marr1976early, marr1980visual}, researchers generally believed that sketches consist of a series of assertions. These assertions are abstract descriptions of visual elements such as edges, lines, and blobs in an image, forming a symbolic representation of the image and laying the theoretical foundation for visual representation research. 
Each assertion represents a certain visual feature in the image: 
1. \textit{Edge} denotes brightness changes (e.g., contours or shadows); 
2.  \textit{Shading-Edge} indicates lighting-induced edges, often defining 3D shape; 
3. \textit{Extended-Edge}: denotes continuous edges, possibly referring to object boundaries or surface features; 
4. \textit{Line}: represents elongated linear features; 
5. \textit{Blob}: represents regional features, such as bright circular or irregularly shaped areas.

Inspired by these cognitive processes, we deconstruct the structural features of the image into three categories: \textbf{single lines}, \textbf{double lines}, and \textbf{soft edges}. These features correspond to the visual levels naturally formed in human perception of an object's appearance. We attempt to apply these features to the sketch-to-image generation task, using them as guiding information that progressively influences the generation process of the diffusion model. By analyzing the input sketch in three levels, prominent features are processed as double lines, the division of object blocks~\cite{hertzmann2020line, li2019photo} and geometric edges as single line visual representations, while discrete low-level visual features soft edges are used to guide spatial gradient representations and subsequent texture generation to accurately reproduce spatial relationships and fit high-frequency texels of the object. 
\subsection{Imprimatura: Layer Design in Painting}
\label{section_3.3}
We draw inspiration from a classical painting technique known as “Imprimatura”~\cite{faragasso2020student, osborne1970oxford}. In this approach, the painting process is divided into two stages: Underpainting and Glazing~\cite{bruyn1982painting}. As shown in Figure~\ref{fig:3_2_Imprimatura}, in art, an underpainting is an initial layer of paint applied to a ground, which serves as a base for subsequent layers of paint~\cite{oudry1752discourse}. Therefore, artists typically use a grayscale tone in this stage to outline contrasts in light and shadow as well as structural form, laying a foundation for later color application. After completing the underpainting, a glaze is a thin transparent or semi-transparent layer on a painting which modifies the appearance of the underlying paint layer~\cite{keith1996giampietrino}. Glazes can change the chroma, value, hue, and texture of a surface~\cite{Glazing1952John}. This staged technique excels by laying a robust foundation of grayscale tones and lighting contrasts, which facilitates the subsequent surface layer coloring. It produces a final image with boasts profound color depth, pronounced dimensionality and real texture.

In image processing, traditional methods such as Total Variation~\cite{rudin1992nonlinear}, Bilateral Filter~\cite{tomasi1998bilateral}, Guided Filter~\cite{he2012guided}, Frequency Decomposition~\cite{gao2021neural}, and Texture Analysis~\cite{liao2017visual} as well as deep learning-based approaches (e.g.,~\cite{chen2017fast, fan2018image, liu2016learning}), aim to remove the content details from the style image, preserving only key color and texture characteristics as style codes. These codes are then combined with specified content to achieve effective domain transfer. However, when applied to object-centered generation tasks, these methods often fall short due to the complex interweaving of different attributes in image formation, resulting in issues like a lack of realistic lighting effects~\cite{StyleID, AesPA-Net} or conflicts between appearance textures and the original sketch details~\cite{crossimageattention}. Therefore, we propose applying the principles of Imprimatura to the sketch-to-image generation task. We designate the Underpainting process as \textbf{Base layer shaping} and Glazing as \textbf{Surface layer coloring}. Base Layer Shaping handles implicit image info like lighting, illumination, and shading reflectance. Features related to texel and color are addressed in the Surface Layer Coloring stage.
\begin{figure*}[t!]
    \centering
    \includegraphics[width=\textwidth]{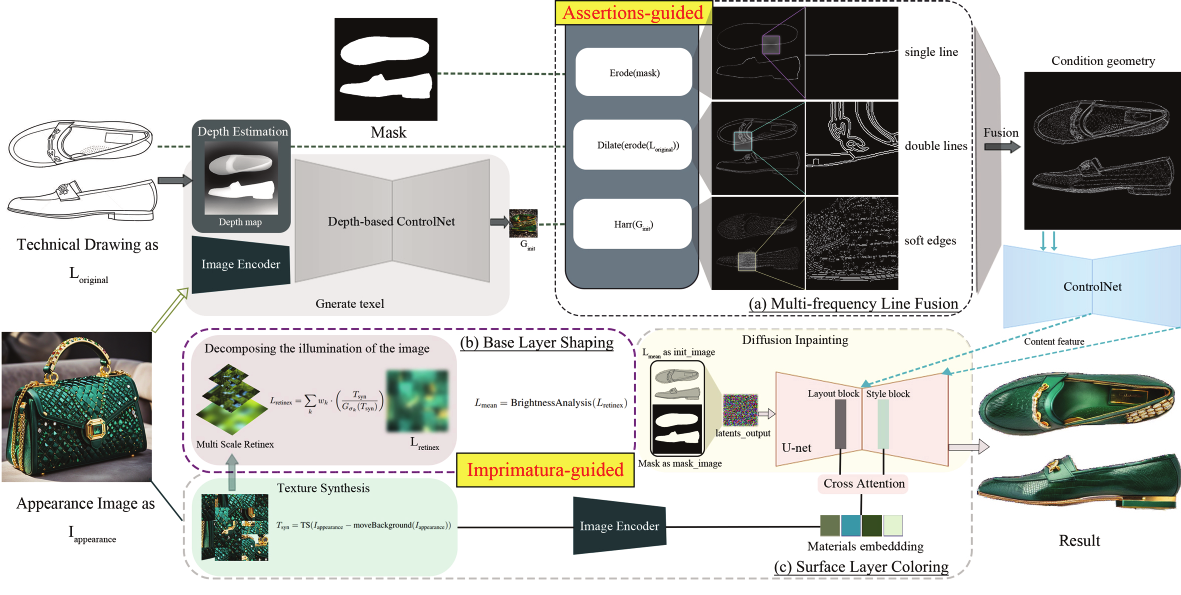}
    \vspace{-6mm}
    \caption{\textbf{Our Workflow.} 
    The process begins with a design drawing $L_{\text{original}}$ and an appearance image $I_{\text{appearance}}$. Depth-based ControlNet estimates depth and generates soft edges to guide the synthesis. 
    (a) The Multi-frequency Line Fusion module employs assertion-guided techniques to enhance structural detail control. 
    (b) Base Layer Shaping decomposes the illumination of the appearance image using a multi-scale retinex approach, generating retinex illumination layers $L_{\text{retinex}}$ to balance brightness. 
    (c) Surface Layer Coloring refines the output by utilizing layout and style blocks in a U-net with cross-attention for accurate material embedding.}
    \vspace{-6mm}
    \label{fig:pipeline}
\end{figure*}
\begin{figure}[t]
    \centering
    \includegraphics[width=\linewidth]{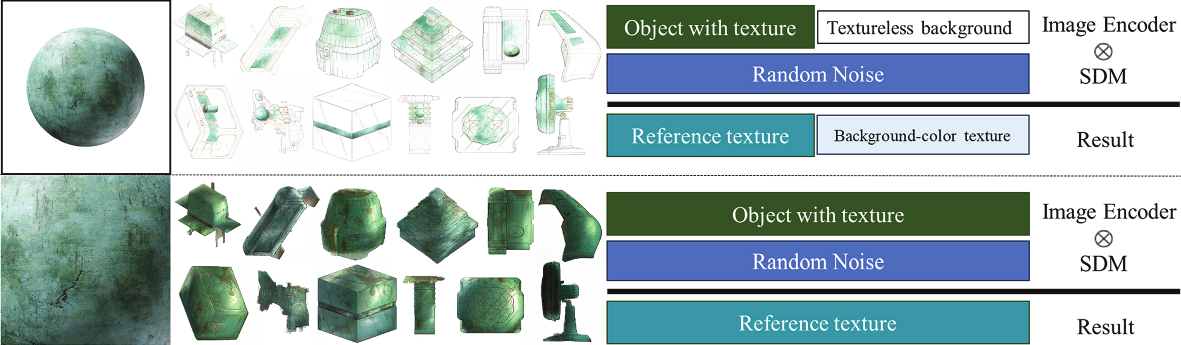}
    \caption{\textbf{Moving Background's Impact on Texture Synthesis.} 
    By removing irrelevant background regions, materials embedding become more accurate materials embedding.}
    \vspace{-6mm}
    \label{fig:moveBG}
\end{figure}

\section{Method}
We propose a visual cognition-guided image generation method from design drawings as shown in Figure~\ref{fig:pipeline}.
The framework is based on the inpainting mode of ControlNet~\cite{ControlNet}, IP-adapter~\cite{IP-Adapter} and Stable Diffusion~\cite{rombach2022high}, embedding human visual cognitive knowledge into network modules for precise control over image generation.
\subsection{Knowleged-guided Structure Preservation}
According to~\ref{section_3.2}, inspired by the hierarchical processing of visual signals in the human brain and the varying attention humans give to brushstrokes while drawing, we propose three-level edge representations: a continuous closed \textit{single line}, a continuous \textit{double line}, and a collection of discrete point lines, namely \textit{soft edges}.
These levels are designed to build a meaningful and structured visual framework for input line drawings. Our design concept is based on our three core insights about the lines. For details of the insights, please see the supplementary materials.

Building on these insights, we apply morphological filtering and frequency analysis to the original line drawing to construct the multi-frequency line fusion module. 
Specifically, for single-line attribution, there are several stroke-level attribution methods (such as those in \cite{2024sketchmean, bohle2022b, selvaraju2017grad}), but we currently opt to use a mask to capture the outer contour, which in our task serves only to differentiate foreground from background. The double-line attribution is represented as $L_{\text{double}} = \text{dilate}(\text{erode}(L_{\text{original}}))$, while the single-line attribution is obtained by applying erosion to the mask as $L_{\text{single}} = \text{erode}(\text{Mask})$.

Next, we use~\cite{yang2024depth} to predict the depth information of the line drawing and generate it for the first time using ControlNet-depth, in combination with the specified appearance image. The initial generation result $G_{\text{initial}}$ is then used to extract soft edges $S_{\text{soft}}$ that capture spatial gradients and texture characteristics as $S_{\text{soft}} = \text{Haar}(G_{\text{initial}})$. These soft edges act as high-frequency constraints in subsequent appearance generation, ensuring accuracy and naturalness of the results in terms of texture primitives.
Finally, the three levels of edges are fused as $Condition_{\text{geometry}} = \text{Fusion}(L_{\text{double}}, L_{\text{single}}, S_{\text{soft}})$ and input into ControlNet as content features. 

\subsection{Knowleged-guided Appearance Transfer}
\begin{figure*}[t]
    \centering
    \vspace{-7mm}
    \includegraphics[width=\textwidth]{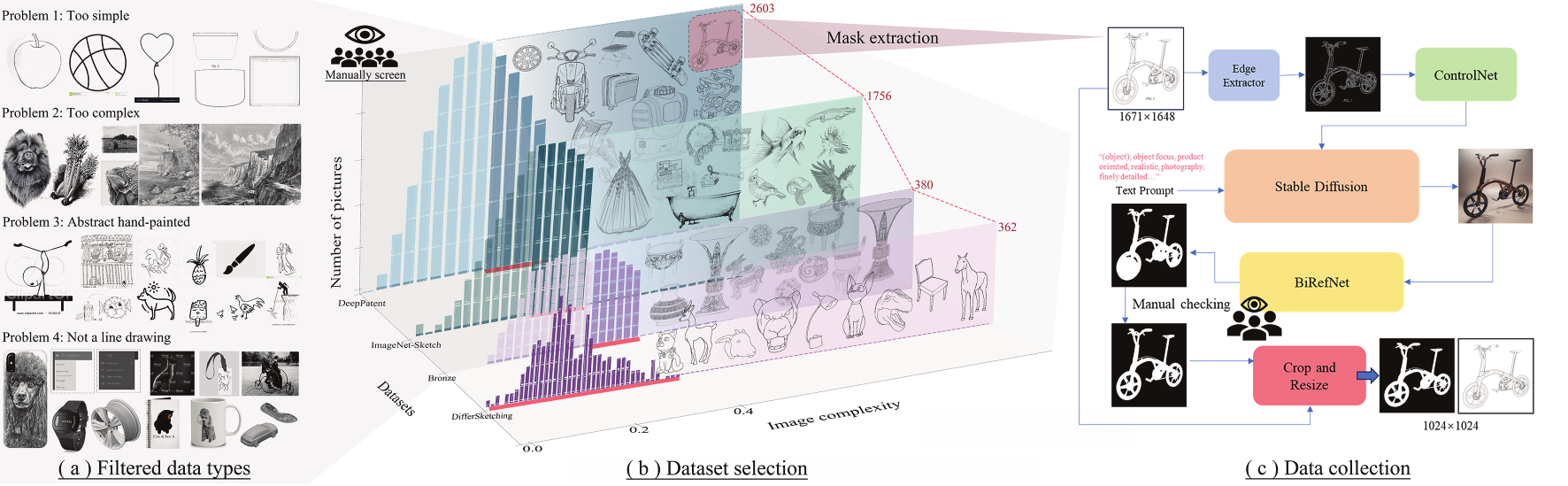}
    \caption{\textbf{Construction of \datasetName Dataset.} 
    (a) shows the types of data we filtered out.
    (b) shows an overview of the data after the initial screening based on image complexity and manual removal of noise data. 
    (c) shows the data preprocessing of the selected data, including the automatic processing process of mask and three rounds of manual verification. After (b)(c) two processes, we obtained 5101 precious line drawings.}
    \vspace{-6mm}
    \label{fig:5_datasetProcess}
\end{figure*}

Accoding to~\ref{section_3.3}, in order to achieve high-fidelity appearance transfer effects, we borrowed \textit{Imprimatura}, a coloring technique in classical oil paintings, and divided the appearance generation process into two stages: Base Layer Shaping and Surface Layer Coloring.

Initially, we observed that the IP-Adapter and ControlNet framework for appearance rendering often encounters color deviations, particularly when the generated image’s color and texture differ from the reference image. 
We analyzed that this issue due to the fact that the material encoding features of the image encoder failed to fully contain the necessary texture and color information. 
Thus, we first remove the background of the reference appearance image, retain only the valid pixel area. Then, inspired by~\cite{noroozi2016unsupervised, he2022masked}, we divide valid pixel area into small patches and reassemble it into a new texture reference image as $T_{\text{syn}} = \text{TS}(I_{\text{appearance}} - \text{moveBackground}(I_{\text{appearance}}))$. As shown in Figure~\ref{fig:moveBG}, this operation can significantly improve the control ability of texture and color. For more discussion on the generation effects of patch size selection, please see the supplementary material.

\begin{figure*}[t]
    \centering
    \vspace{-3mm}
    \includegraphics[width=\textwidth]{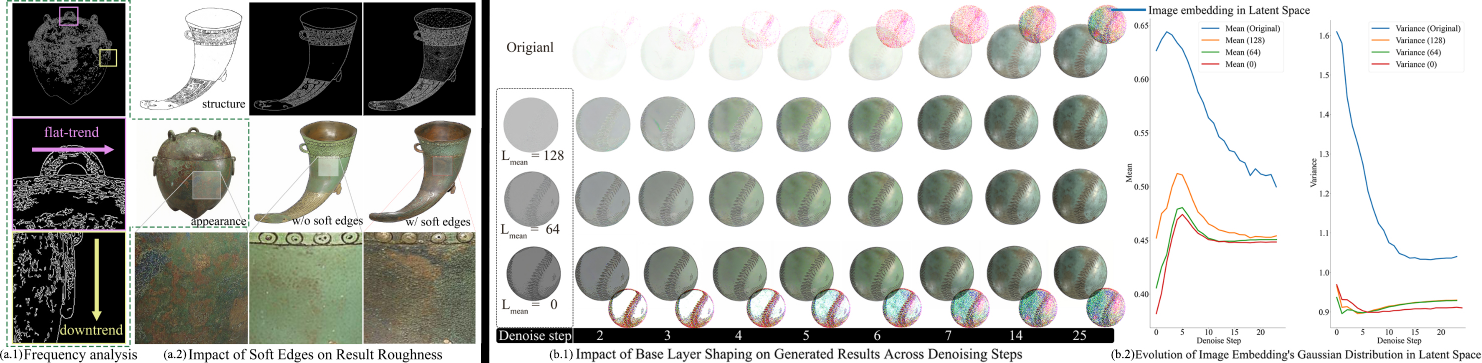}
    \caption{(a.1) Frequency Analysis of Texture Elements in Real Photos. 
    (a.2)Improved material fidelity through \textit{soft edge} constraints.
    (b.1)shows the progressive influence of base layer shaping at different mean latent levels ($L_{\text{mean}} = 0, 64, 128$) on the generated result as denoising steps advance. 
    (b.2) Illustration of the evolution of mean and variance in the Gaussian distribution of image embeddings within latent space. Compared to the original, the data's iterative direction remains consistent, but the overall distribution shifts.
    }
    \vspace{-3mm}
    \label{fig:softGray}
\end{figure*}

After optimizing the appearance image, we proceed to the \textit{Base Layer Shaping} stage. Here, we use a carefully designed grayscale distribution from the reference image as the starting point for light and shadow generation, guiding the denoising process. By affecting the mean and variance of the initial noise distribution, we can control the overall brightness of the generated result (as Figure~\ref{fig:softGray}(b)). 
We use a multi-scale Retinex model to extract the lighting properties of the appearance image. By applying Gaussian blur at different scales, we capture the lighting changes of different spatial frequencies, and perform normalization and weighted combination. The structure obtained by Retinex is then subjected to brightness analysis as $L_{\text{mean}}, \sigma^2_L = \text{BrightnessAnalysis}(L_{\text{retinex}})$
The results of the brightness analysis are then used as the initial starting point for the diffusion model denoising ($x_0' = L_{\text{mean}} + (x_0 - L_{\text{mean}}) \times 0.5$). Thus, we establish a soft link between the lighting properties of the generated image and the reference image as $\hat{x}_t = \sqrt{\bar{\alpha}_t} x_0' + \sqrt{1 - \bar{\alpha}_t} z_t$.

In the \textit{Surface Layer Coloring} stage, we use decoupled cross-attention mechanism to integrate text features, content features, and appearance features.
The following framework includes a small projection network that maps the image embedding to a feature sequence (N=4) matching the pre-trained diffusion model’s text features. The text embedding is injected into Unet via cross-attention, with ControlNet mapping the semantic layout. Material embedding from appearance image is injected into specific attention layers (up blocks.0.attentions.1 for color, material, and atmosphere while down blocks.2.attentions.1 for structure and spatial composition), addressing content and style fusion differences~\cite{Instantstyle}. 
Thus, the synthesized texture image directs appearance generation, with soft edges acting as high-frequency guides, ensuring alignment with texture patterns and promoting natural layout changes.
\section{\datasetName Dataset}
\noindent \textbf{Knowledge-guided data collection:} Previous research on sketches has not clearly defined and classified them. Specifically, although there are currently multiple manually drawn sketch datasets, they are usually too abstract and unverified (such as Sketchy~\cite{sangkloy2016sketchy}, QuickDraw~\cite{jongejan2016quick}, Pseudosketches~\cite{COGS_ECCV2022}, TU-Berlin~\cite{eitz2012humans}, QMUL-Shoe/Chair~\cite{song2018learning}, and OpenSketches~\cite{OpenSketch19}). Therefore, we quantify and classify sketches based on image complexity (IC). We used ICNet~\cite{feng2022ic9600} to quantify the IC of line drawings, scoring each image between 0 and 1. Every drawing was sorted by IC, and we focused on selecting those in the central threshold range, avoiding noisy data (Figure~\ref{fig:5_datasetProcess}(a)). Then six CV/CG professionals manually screened the images, removing those that did not meet our task’s criteria.
As shown in Figure~\ref{fig:5_datasetProcess}(b), we designed a data collection pipeline and finally collected 5101 line drawings from four design datasets: Bronze, DifferSketching~\cite{xiao2022differsketching}, ImageNet-Sketch~\cite{wang2019learning}, DeepPatent~\cite{kucer2022deeppatent}. The screening threshold interval and number of images of each dataset were as follows:
Bronze ($0.2576 \sim 0.2903$, $380$ images); DifferSketching ($0.0461 \sim 0.2165$, $362$ images); ImageNet-Sketch ($0.2500 \sim 0.2650$, $1756$ images); DeepPatent ($0.2715 \sim 0.2790$, $2603$ images).

\noindent \textbf{Data preprocessing:} As shown in Figure~\ref{fig:5_datasetProcess}(c), after screening the ideal line drawing data for complex images, we built an automated workflow for line drawing mask extraction, background masking, and size cropping for subsequent generation and evaluation experiments. 

We provide more detailed instructions on constructing the dataset, such as the threshold selection process, noisy data definition, and manual verification principles in the supplementary materials.

\begin{figure*}[t]
    \captionsetup{type=figure}
    \includegraphics[width=\textwidth]{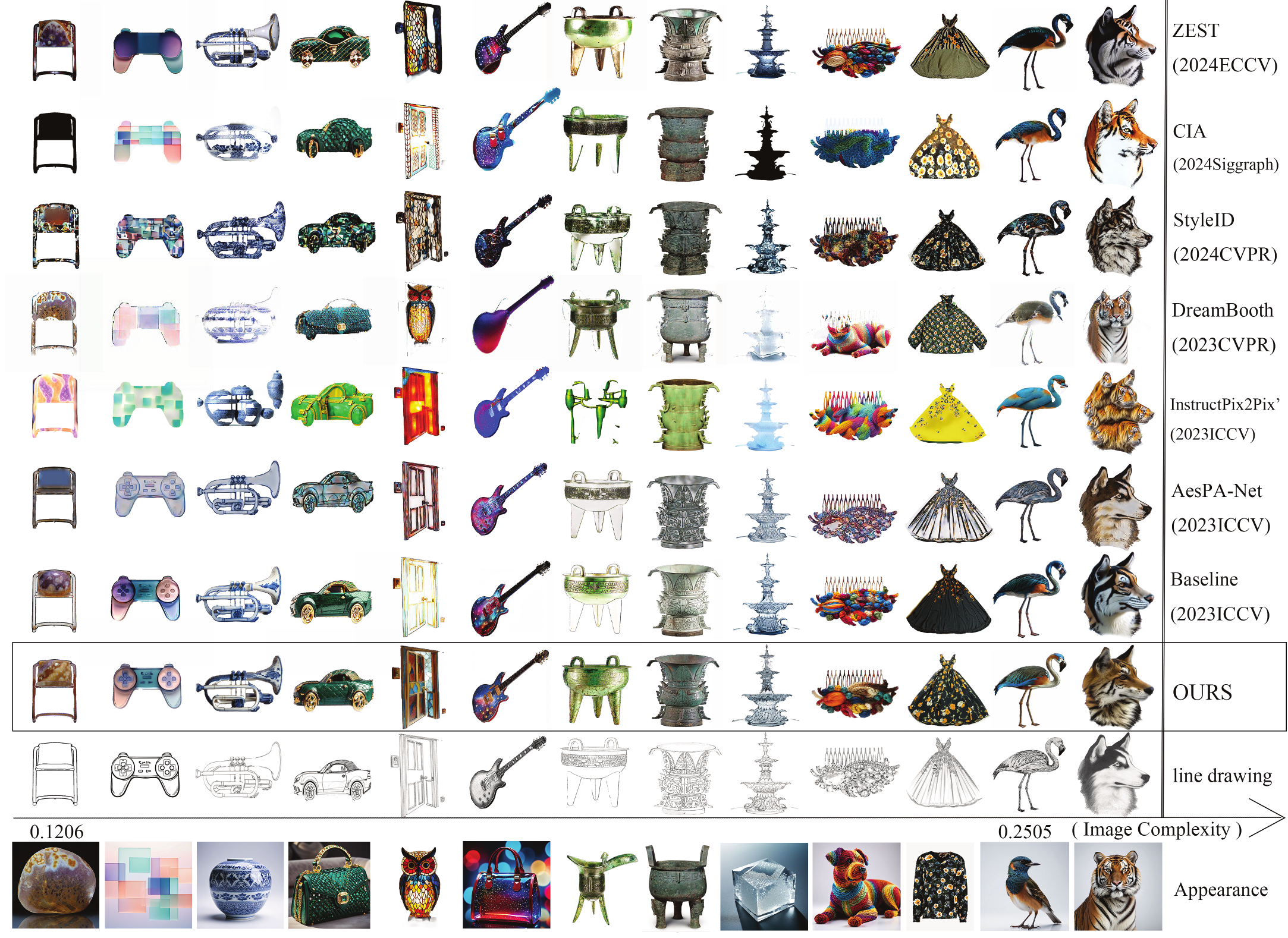}
    \captionof{figure}{
    \textbf{Our work is compared with other SOTAs in the qualitative results of processing technical drawings with fine structures and images of specified materials. }As can be seen from the figure, we can generate more accurate textures while maintaining finer structures (such as the detailed patterns of bronze vessels in the 7th and 8th columns), and have more appropriate color representation for material images (such as chairs in the 1st column, hairpins in the 10th column, and skirts in the 11th column). In addition, our model also shows stronger generalization ability and robustness when facing transfer tasks without corresponding relationships, and will not collapse (such as domain transfer from tiger to dog in the 13th column).
    }
    \vspace{-3mm}
    \label{fig:quailitativeResult}
\end{figure*}

\begin{table*}[t]
\centering
\captionof{table}{
\textbf{Quantitative evaluation of different methods on \datasetName datasets.} We evaluate the quality of results across three dimensions using eight metrics. The results demonstrate that our method strikes a good balance between perceptual quality and accuracy, especially in preserving image details and textures. The best value is highlighted in \textcolor{black}{\colorbox[HTML]{FFCCC9}{red}}, and the second-best in \textcolor{black}{\colorbox[HTML]{FFFFC7}{yellow}}.
}
\scalebox{0.85}{
\begin{tabular}{c|c|ccccc|ccccc|ccc}
\hline
 &  &  & \multicolumn{3}{c}{\textbf{Overall quality}} &  &  & \multicolumn{3}{c}{\textbf{Appearance perception}} &  &  & \multicolumn{2}{c}{\textbf{Edge-fidelity}} \\ \cline{3-15} 
\multirow{-2}{*}{\textbf{Dataset}} & \multirow{-2}{*}{\textbf{Method}} &  & FID\(\downarrow\) & LPIPS\(\downarrow\) & CLIP$_\text{i}$\(\uparrow\) &  &  & PSNR\(\uparrow\) & CH\(\downarrow\) & GLCM\(\downarrow\) &  &  & SSIM\(\uparrow\) & CD\(\downarrow\) \\ \hline
\textbf{} & \textbf{Ours} &  & \cellcolor[HTML]{FFCCC9}{\ul \textbf{86.31}} & \cellcolor[HTML]{FFCCC9}{\ul \textbf{0.25}} & \cellcolor[HTML]{FFCCC9}{\ul \textbf{0.86}} &  &  & 25.20 & \cellcolor[HTML]{FFCCC9}{\ul \textbf{0.20}} & \cellcolor[HTML]{FFCCC9}{\ul \textbf{2.42}} &  &  & \cellcolor[HTML]{FFCCC9}{\ul \textbf{0.83}} & \cellcolor[HTML]{FFCCC9}{\ul \textbf{2.38}} \\
 & Baseline~\cite{ControlNet,IP-Adapter} &  & \cellcolor[HTML]{FFFFC7}88.26 & \cellcolor[HTML]{FFFFC7}0.35 & \cellcolor[HTML]{FFFFC7}0.84 &  &  & 24.48 & 0.23 & 29.23 &  &  & \cellcolor[HTML]{FFFFC7}0.81 & \cellcolor[HTML]{FFFFC7}2.72 \\
\textbf{Bronze} & CIA~\cite{crossimageattention} &  & 140.18 & 0.47 & 0.75 &  &  & 20.93 & 0.68 & \cellcolor[HTML]{FFFFC7}2.45 &  &  & 0.73 & 263.97 \\
IC:0.2576-0.2903 & ZeST~\cite{ZeST} &  & 113.79 & 0.41 & 0.78 &  &  & \cellcolor[HTML]{FFFFC7}25.87 & \cellcolor[HTML]{FFFFC7}0.21 & 5.71 &  &  & 0.72 & 6.32 \\
 & DreamBooth~\cite{Dreambooth} &  & 113.36 & 0.38 & 0.81 &  &  & \cellcolor[HTML]{FFCCC9}{\ul \textbf{27.15}} & 0.25 & 11.23 &  &  & 0.77 & 9.76 \\
 & InstructPix2Pix'~\cite{brooks2023instructpix2pix, IP-Adapter} &  & 223.72 & 0.48 & 0.74 &  &  & 24.15 & 0.29 & 36.72 &  &  & 0.79 & 15.51 \\ \hline
 & \textbf{Ours} &  & \cellcolor[HTML]{FFFFC7}211.37 & \cellcolor[HTML]{FFCCC9}{\ul \textbf{0.23}} & \cellcolor[HTML]{FFCCC9}{\ul \textbf{0.83}} &  &  & \cellcolor[HTML]{FFFFC7}19.25 & \cellcolor[HTML]{FFCCC9}{\ul \textbf{0.62}} & \cellcolor[HTML]{FFCCC9}{\ul \textbf{1.24}} &  &  & \cellcolor[HTML]{FFCCC9}{\ul \textbf{0.96}} & \cellcolor[HTML]{FFCCC9}{\ul \textbf{2.48}} \\
 & Baseline~\cite{ControlNet,IP-Adapter} &  & 218.83 & \cellcolor[HTML]{FFFFC7}0.24 & \cellcolor[HTML]{FFFFC7}0.81 &  &  & 19.11 & 0.69 & 30.75 &  &  & \cellcolor[HTML]{FFFFC7}0.95 & \cellcolor[HTML]{FFFFC7}3.94 \\
\textbf{DifferSketching} & CIA~\cite{crossimageattention} &  & 237.38 & 0.47 & 0.68 &  &  & \cellcolor[HTML]{FFCCC9}{\ul \textbf{20.16}} & \cellcolor[HTML]{FFFFC7}0.64 & \cellcolor[HTML]{FFFFC7}1.61 &  &  & 0.91 & 241.15 \\
IC:0.0461-0.2165 & ZeST~\cite{ZeST} &  & 242.88 & 0.25 & 0.78 &  &  & 19.24 & 0.68 & 2.26 &  &  & 0.94 & 15.35 \\
 & DreamBooth~\cite{Dreambooth} &  & \cellcolor[HTML]{FFCCC9}{\ul \textbf{202.52}} & 0.24 & 0.77 &  &  & 18.88 & 0.76 & 37.48 &  &  & 0.95 & 30.93 \\
 & InstructPix2Pix'~\cite{brooks2023instructpix2pix, IP-Adapter} &  & 235.87 & 0.29 & 0.76 &  &  & 19.15 & 0.74 & 30.71 &  &  & 0.94 & 10.32 \\ \hline
 & \textbf{Ours} &  & \cellcolor[HTML]{FFCCC9}{\ul \textbf{100.71}} & \cellcolor[HTML]{FFCCC9}{\ul \textbf{0.20}} & \cellcolor[HTML]{FFCCC9}{\ul \textbf{0.86}} &  &  & \cellcolor[HTML]{FFCCC9}{\ul \textbf{20.93}} & \cellcolor[HTML]{FFCCC9}{\ul \textbf{0.33}} & \cellcolor[HTML]{FFCCC9}{\ul \textbf{3.19}} &  &  & \cellcolor[HTML]{FFCCC9}{\ul \textbf{0.88}} & \cellcolor[HTML]{FFCCC9}{\ul \textbf{5.89}} \\
 & Baseline~\cite{ControlNet,IP-Adapter} &  & \cellcolor[HTML]{FFFFC7}111.04 & \cellcolor[HTML]{FFFFC7}0.23 & \cellcolor[HTML]{FFFFC7}0.85 &  &  & 19.85 & \cellcolor[HTML]{FFFFC7}0.36 & 28.37 &  &  & \cellcolor[HTML]{FFFFC7}0.85 & \cellcolor[HTML]{FFFFC7}7.10 \\
\textbf{ImageNet\_Sketch} & CIA~\cite{crossimageattention} &  & 205.18 & 0.47 & 0.66 &  &  & 20.92 & 0.50 & \cellcolor[HTML]{FFFFC7}16.03 &  &  & 0.80 & 240.64 \\
IC:0.2500-0.2650 & ZeST~\cite{ZeST} &  & 167.78 & 0.29 & 0.79 &  &  & \cellcolor[HTML]{FFFFC7}20.22 & 0.37 & 28.65 &  &  & 0.83 & 12.98 \\
 & DreamBooth~\cite{Dreambooth} &  & 198.66 & 0.32 & 0.77 &  &  & 18.88 & 0.39 & 34.79 &  &  & 0.85 & 33.94 \\
 & InstructPix2Pix'~\cite{brooks2023instructpix2pix, IP-Adapter} &  & 127.03 & 0.25 & 0.82 &  &  & 19.15 & 0.40 & 28.62 &  &  & 0.84 & 14.00 \\ \hline
 & \textbf{Ours} &  & \cellcolor[HTML]{FFCCC9}{\ul \textbf{107.23}} & \cellcolor[HTML]{FFCCC9}{\ul \textbf{0.30}} & \cellcolor[HTML]{FFCCC9}{\ul \textbf{0.84}} &  &  & \cellcolor[HTML]{FFCCC9}{\ul \textbf{28.86}} & \cellcolor[HTML]{FFCCC9}{\ul \textbf{0.25}} & \cellcolor[HTML]{FFCCC9}{\ul \textbf{3.42}} &  &  & \cellcolor[HTML]{FFCCC9}{\ul \textbf{0.83}} & \cellcolor[HTML]{FFCCC9}{\ul \textbf{8.12}} \\
 & Baseline~\cite{ControlNet,IP-Adapter} &  & \cellcolor[HTML]{FFFFC7}130.32 & \cellcolor[HTML]{FFFFC7}0.39 & \cellcolor[HTML]{FFFFC7}0.77 &  &  & 20.31 & 0.38 & 20.25 &  &  & \cellcolor[HTML]{FFFFC7}0.80 & \cellcolor[HTML]{FFFFC7}14.49 \\
\textbf{DeepPatent} & CIA~\cite{crossimageattention}&  & 250.19 & 0.58 & 0.65 &  &  & \cellcolor[HTML]{FFFFC7}21.39 & 0.47 & \cellcolor[HTML]{FFFFC7}13.34 &  &  & 0.72 & 224.91 \\
IC:0.2715-0.2790 & ZeST~\cite{ZeST} &  & 182.03 & 0.48 & 0.73 &  &  & 20.22 & 0.37 & 26.61 &  &  & 0.76 & 27.56 \\
 & DreamBooth~\cite{Dreambooth} &  & 159.45 & 0.44 & 0.73 &  &  & 19.74 & \cellcolor[HTML]{FFFFC7}0.35 & 31.53 &  &  & 0.77 & 46.55 \\
 & InstructPix2Pix'~\cite{brooks2023instructpix2pix, IP-Adapter} &  & 179.62 & 0.44 & 0.75 &  &  & 20.29 & 0.42 & 24.94 &  &  & 0.78 & 28.03 \\ \hline
\end{tabular}
}
\label{table:quantitativeResult}
\vspace{-6mm}
\end{table*}

\section{Experiments}

\subsection{Evaluation Metrics and Comparison Models}
\noindent \textbf{Metrics:} We evaluate the generated results from three dimensions: edge fidelity, appearance transfer quality, and overall perceptual quality. In terms of edge fidelity, we use two metrics: structural similarity index (SSIM) and chamfer distance (CD). About the quality of appearance transfer, we introduce three metrics: gray-level co-occurrence matrix (GLCM), peak signal-to-noise ratio (PSNR)~\cite{riba2020kornia}, and color histogram (CH) loss. 
About overall perceptual quality, we used Frechet Inception Distance (FID), Learned Perceptual Patch Similarity (LPIPS)~\cite{zhang2018perceptual}, and Contrastive Language Image Pre-training (CLIP). For detailed description of the metrics, please see the supplementary material.

\noindent \textbf{Comparison Models:}
We selected these sota works for experimental results evaluation. 
ZeST~\cite{ZeST}, 
Cross-Image Attention (CIA)~\cite{crossimageattention}, 
StyleID~\cite{StyleID}, 
DreamBooth~\cite{Dreambooth}, 
InstructPix2Pix~\cite{brooks2023instructpix2pix} + IP-Adapter (InstructPix2Pix'), 
AesPA-Net~\cite{AesPA-Net}, 
ControNet-Canny + IP-Adapter (Baseline). Please see the supplementary material for the relevant model versions and specific experimental setup details.

\subsection{Quantitative and Qualitative Analysis}
As shown in Table~\ref{table:quantitativeResult}, we conducted a comprehensive quantitative experiment. The results show that our method achieves the best performance in most of the metrics. We also conducted a qualitative experiment comparing with seven SOTAs. As shown in Figure~\ref{fig:quailitativeResult}, CIA, DreamBooth, and InstructPix2Pix’ struggle to maintain the correct line drawing structure. StyleID and AesPA-Net lost the maintenance of light and shadow and realism during the transfer process. When comparing our approach with ZeST and Baseline, it is evident that ZeST lacks effectiveness in preserving line drawing details. Baseline, meanwhile, some of the generated results lack realism and show ineffective material transfer. Considering the results of the above experiments, our method outperforms others in preserving the structure and details of line drawings, seamlessly transferring materials from appearance images, and generating images with enhanced realism and visual quality.

\subsection{User Study and Ablation Study} 
To further validate the authenticity of our generated results from the user's perspective, we conducted a user study with 20 participants. Users were asked to rank results from six methods based on three evaluation criteria: edge fidelity, appearance perception, and overall perception. Participants ranked nine sets of randomly provided examples according to these criteria. We achieved the highest scores in all three dimensions as shown in Figure~\ref{fig:userStudy}. We provide more details of the user study in the supplementary material.

The ablation study, presented in the supplementary materials, evaluates the effectiveness of key design choices in our method across four aspects: (1) the role of double lines in controlling pattern thickness; (2) soft edges as high-frequency constraints to enhance texture generation; (3) the impact of surface layer coloring; and (4) the influence of patch size in texture synthesis. 
\begin{figure*}[ht]
    \centering
    \vspace{-6mm}
    \includegraphics[width=\textwidth]{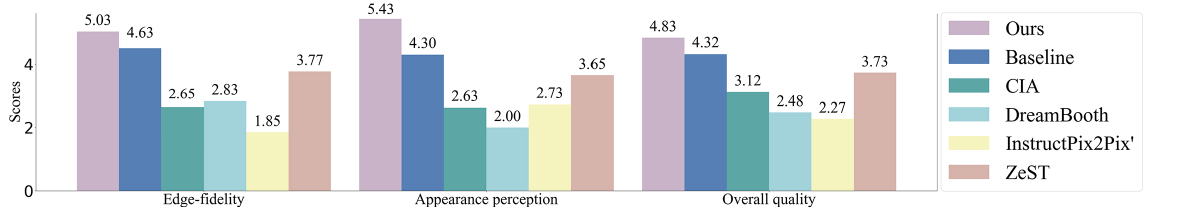}
    \caption{Result of User Study.}
    \vspace{-6mm}
    \label{fig:userStudy}
\end{figure*}

\section{Conclusion}
Our work represents an innovative attempt to integrate human drawing knowledge into the image generation process, advancing the understanding and creation of design drawings. This methodology deepens existing artistic knowledge and contributes to the evolution of digital art, offering new perspectives and approaches for generating professional drawings.
Our work also has several limitations. We struggled with line drawings that contain excessive shadows or complex textures, as well as tasks requiring precise semantic correspondence. For details on the failure examples and demonstrations, please see the supplementary materials.

{
    \small
    \bibliographystyle{ieeenat_fullname}
    \bibliography{main}
}
\end{document}